\documentclass[cameraready]{Interspeech}

\usepackage{cite}
\usepackage{microtype}

\title{ZeroSyl: Simple Zero-Resource Syllable Tokenization \\ for Spoken Language Modeling}

\author{Nicol}{Visser}
\author{Simon}{Malan}
\author{Danel}{Slabbert}
\author{Herman}{Kamper}

\address{Electrical and Electronic Engineering, Stellenbosch University, South Africa}

\email{\{16986431, 24227013, 24051055, kamperh\}@sun.ac.za}

\usepackage{tabularx}
\usepackage{multirow}

\newcolumntype{C}{>{\centering\arraybackslash}X}
\newcolumntype{L}{>{\raggedright\arraybackslash}X}
\newcolumntype{R}{>{\raggedleft\arraybackslash}X}

\usepackage{stfloats}

\usepackage{xcolor}

\usepackage[table]{xcolor}

\begin{document}

\maketitle

\begin{abstract}
Pure speech language models aim to learn language directly from raw audio without textual resources.
A key challenge is that discrete tokens from self-supervised speech encoders result in excessively long sequences, motivating recent work on syllable-like units.
However, methods like Sylber and SyllableLM rely on intricate multi-stage training pipelines.
We propose ZeroSyl, a simple training-free method to extract syllable boundaries and embeddings directly from a frozen WavLM model.
Using L2 norms of features in WavLM's intermediate layers, ZeroSyl achieves competitive syllable segmentation performance.
The resulting segments are mean-pooled, discretized using K-means, and used to train a language model.
ZeroSyl outperforms prior syllabic tokenizers across lexical, syntactic, and narrative benchmarks.
Scaling experiments show that while finer-grained units are beneficial for lexical tasks, our discovered syllabic units exhibit better scaling behavior for syntactic modeling.
\end{abstract}

\section{Introduction}

Advances in self-supervised learning (SSL)~\cite{mohamed2022sslreview} have made it possible to train language models directly on audio data:
instead of using tokenized text, a causal language model is trained on tokens derived from an SSL model~\cite{dunbar2022lessons}.
Without relying on any textual resources, this framework would enable natural language processing for languages lacking the massive speech-text datasets required by traditional speech-driven technology.
These pure speech language models showed early potential \cite{lakhotia2021gslm}, but subsequent progress has slowed.

\begin{figure}[t!]
  \centering
  \includegraphics[width=0.99\linewidth]{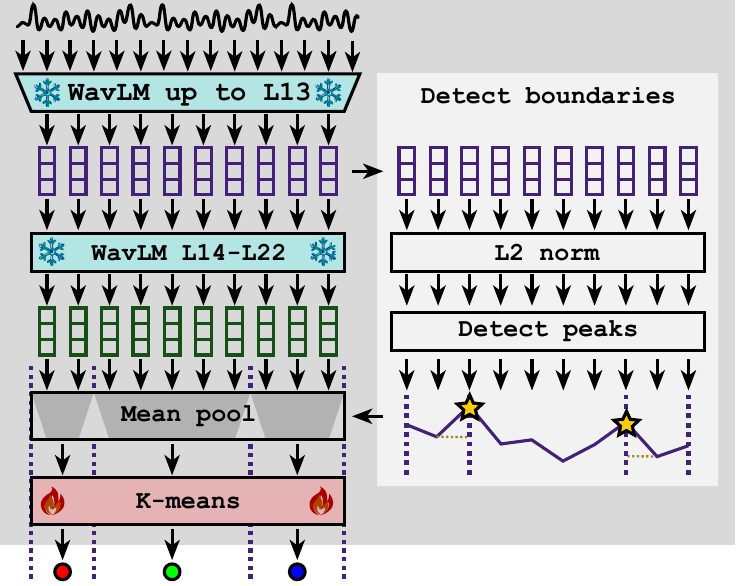}
  \vspace*{-2.5pt}
  \caption{
    ZeroSyl detects syllabic boundaries using prominence-based peak detection on features from layer 13 of a frozen WavLM. It then mean pools semantic features from layer 22 within the discovered boundaries and clusters these using spherical $K$-means. A language model is trained on these cluster~IDs.
  }
  \vspace*{-2pt}
  \label{fig:pipeline}
\end{figure}

The decline in progress is most evident in syntactic modeling, where performance has plateaued since 2023.
AudioLM~\cite{borsos2023audiolm} remains the state-of-the-art, yet it relies on a resource-intensive framework:
a high-capacity language model trained on high-bitrate tokens from w2v-BERT~XL~\cite{chung2021w2vbert}.
Moreover, recent work indicates that pure speech modeling scales less favorably than text~\cite{cuervo2023scaling}, suggesting that simply increasing model size and data volume will not solve the problem.
This has resulted in a shift toward discovering more effective speech units in order
to close the performance gap with textual models of comparable size~\cite{poli2025spidr}.

One hypothesis is that the fine granularity of standard speech tokens creates a bottleneck~\cite{algayres2023wordsized}.
Because discrete speech tokens are extracted at a much higher rate than text tokens, the resulting sequences are substantially longer, making it difficult to model long-range dependencies.
To mitigate this, studies have attempted to utilize larger phone-, syllable-, or word-like units~\cite{algayres2023wordsized,visser25dpslm,cho2025sylber,baade2025syllablelm}.
Syllable-based modeling has proven to be particularly successful~\cite{cho2025sylber,baade2025syllablelm}.
However, existing syllabic tokenizers rely on complex multi-stage pipelines where an
SSL model must be fine-tuned with a specialized objective.

In this paper, we propose an unsupervised approach to syllable tokenization that does not require the complex methods of previous work.
Figure~\ref{fig:pipeline} shows our simple pipeline.
We find that high-quality syllabic boundaries can be extracted by simply applying peak detection to the L2 norms of frozen WavLM~\cite{chen2022wavlm} features.
We then mean-pool and cluster features within these boundaries to generate a discrete vocabulary.
A symbolic language model~\cite{zhang2022opt} is trained on the resulting sequences.
This approach, which we call ZeroSyl, surpasses the syntactic (sBLIMP) performance of comparable syllabic models and achieves superior performance on the Topic StoryCloze (tSC) narrative benchmark.
It also outperforms syllabic systems on the lexical sWUGGY benchmark.
But in scaling experiments, it does not surpass the performance of a system utilizing fine-grained units~\cite{poli2025spidr}.
Nevertheless, our results demonstrate that high-quality syllable discovery is possible without complex multi-stage training pipelines, paving a simple and effective path for future research in spoken language modeling.
Models and code available at: {\footnotesize{\texttt{https://github.com/nicolvisser/ZeroSyl/}}}

\section{Related work}
\label{sec:related_work}

This work focuses on \textit{pure} speech language models, where text is not used or incorporated at all~\cite{arora2025landscape}.
Standard benchmarks have been developed that assess how well these models capture lexical, syntactic, and semantic properties~\cite{dunbar2021zrc,hassid2024twist}.

\newpage

To date, AudioLM~\cite{borsos2023audiolm} achieves state-of-the-art performance in syntactic modeling.
However, since it uses a 25\,Hz-tokenizer that produces long sequences, a large language model (300M parameters) is subsequently required to capture long-range dependencies.
SpidR~\cite{poli2025spidr} is a recent SSL-based approach that is more lightweight.
It uses online clustering and a novel training objective to improve the SSL representations and resulting speech units.
While SpidR achieves the best lexical modeling scores, its units remain close to the frame level.
As with \mbox{AudioLM}, the sequence lengths are therefore very long, and SpidR has not surpassed AudioLM's syntactic scores.

Instead of directly discretizing SSL features, two noteworthy approaches have turned to syllable units to deal with the long sequence lengths.
Sylber~\cite{cho2025sylber} utilizes a sentence-level distillation bjective~\cite{cho2024sdhubert} to finetune HuBERT~\cite{hsu2021hubert}.
After distillation, syllable boundaries are accessible using cosine similarity between the framewise features.
SyllableLM~\cite{baade2025syllablelm} identifies boundaries by analyzing the HuBERT loss objective.
The idea is that fully masking a syllable results in a distinct disruption of the loss compared to partial masking.
Syllable boundaries are predicted by monitoring the loss as a masked interval sweeps across the utterance.
To reduce computational load at inference time, both approaches distill the boundary information: they train an encoder to predict piecewise-constant targets, derived by mean-pooling the teacher model's embeddings over the discovered segments.
Both approaches therefore require multiple training stages with specially designed architectures and~objectives.

The syllabic methodology shows clear promise for spoken language modeling, but we want to see if a simpler approach is possible.
For this, we build on a simple approach for unsupervised \textit{word} discovery~\cite{pasad2023words}:
prominent peaks in a dissimilarity curve (typically cosine distance) between adjacent features~\cite{bosch2007computational} from a frozen SSL model are taken to indicate word boundaries.
So far, this approach has been optimized for lexical units~\cite{pasad2023words,malan2024unsupervised} and underperforms when used directly to find syllabic units~\cite{malan2025topdownclusteringaffectboundaries}.

\section{Models and methodology}

Our work builds on prominence-based segmentation~\cite{pasad2023words}.
However, where previous work used the cosine similarity between adjacent frames to find word boundaries, we find that the L2 norm of the hidden representations themselves provides a signal corresponding to syllabic structure.
In contrast to Sylber~\cite{cho2025sylber} and SyllableLM~\cite{baade2025syllablelm}, our method is much simpler and does not require further training of the SSL encoder.

\subsection{Boundary detection}
\label{sec:boundaries}

For a given speech utterance, we extract framewise embeddings $H_{1:T}^{[13]}~= [h^{[13]}_1, h^{[13]}_2, \dots, h^{[13]}_T]$ from layer 13 of WavLM Large~\cite{chen2022wavlm}.
We compute the L2 norm for each frame, $n_t = \lVert h^{[13]}_t \rVert_2 $\,, and apply a 3-point moving average filter to reduce high-frequency noise.
We then apply prominence-based peak detection on the smoothed signal:
a peak is identified as a boundary if it has a prominence of at least $\delta = 0.45\sigma$, where $\sigma$ is the standard deviation of the signal $n_t$.

\begin{figure}[t]
  \centering
  \includegraphics[width=0.99\linewidth]{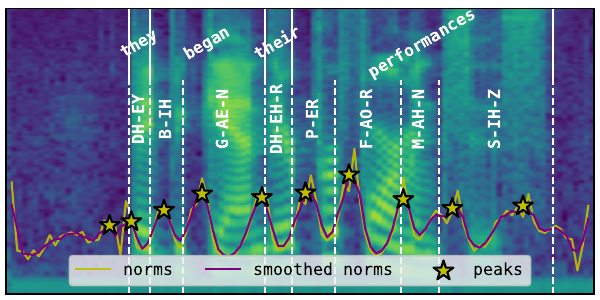}
  \vspace*{-2.5pt}
  \caption{Peak detection on the smoothed L2 norms of framewise features gives our predicted syllable boundaries. The dashed lines indicate the ground truth syllables.}
  \label{fig:zerosyl-boundaries-example}
  \vspace{-10pt}
\end{figure}

Example predictions are shown in Figure~\ref{fig:zerosyl-boundaries-example}.
R-value and token-F1 scores on development data (described later) are best using layer 13, a filter length of 3, and a prominence of $0.45\sigma$.

\subsection{Tokenization}
\label{sec:silence_collapsing}

Once the boundaries $\left[b_1, b_2, ..., b_N\right]$ are established, we extract semantic features from layer 22 of WavLM Large and apply mean-pooling within each segment:
{
\setlength{\abovedisplayskip}{6pt}
\setlength{\belowdisplayskip}{6pt}
\begin{equation}
    e_j = \text{MeanPool}(H^{[22]}_{b_j : b_{j+1}})
\end{equation}
}

\noindent We select this layer because higher WavLM layers are known to capture richer semantic information than those used for boundary detection~\cite{pasad2023comparative}.
Empirically, we find that layer 22 yields the highest mutual information with syllable labels after clustering.

The resulting embeddings $e_j$ are clustered using spherical $K$-means with $K$-means++ initialization.
The $K$-means model is trained until convergence on 100 hours of LibriSpeech~\cite{panayotov2015librispeech} using the \texttt{faiss} library~\cite{douze2024faiss}.
We set the vocabulary size to $K=\text{10\,000}$ following Sylber~\cite{cho2025sylber}, which reports this as the optimal size for syntactic modeling.

After clustering, multiple centroids correspond to silences.
We find that language modeling performance improves when these entries are collapsed to a single vocabulary item (see Table~\ref{tbl:comparative_discovry_and_lm_results} below).
We do this in an unsupervised manner by performing hierarchical clustering on the centroids.
From informal inspection, we know that the two main branches in agglomerative hierarchical clustering correspond to silences and non-silences, respectively.
We pick the smaller branch, which represents silences, and map these items to one vocabulary item.
This reduces the vocabulary size from 10\,000 to 9\,116.

\subsection{Language modeling}
\label{sec:lm_details}

We train a causal language model on the discovered syllabic units.
Following prior work~\cite{baade2025syllablelm, poli2025spidr}, we use the OPT-125M architecture~\cite{zhang2022opt} with a batch size of 81\,920 tokens, and a context length of 2\,048.
The learning rate increases linearly from $0$ to $2\cdot10^{-4}$ for the first 8\% of steps followed by a cosine annealing schedule.
For our comparative experiments in Section~\ref{sec:experiments}, we train the ZeroSyl language model on a single GPU on 6k hours of Libri-Light~\cite{librilight2020} for 25k steps.
For our final scaled experiments in Section~\ref{sec:scaled}, we train on the full 60k hours for 200k steps.

\subsection{Implementation of previous systems}
\label{sec:baselines}

For our comparisons against Sylber~\cite{cho2025sylber} and SyllableLM~\cite{baade2025syllablelm}, we strictly follow the code implementations released by the authors.
Since Sylber did not release a clustering model, we train a spherical $K$-means model with $K=10\,000$ on their embeddings.
As per their specifications~\cite{cho2025sylber}, we include an explicit silence token when more than seven frames correspond to silence.
For SyllableLM, we utilize the official extraction pipeline, including agglomerative clustering and deduplication.
SyllableLM has multiple variants.
To ensure the strongest possible baseline, we consider all configurations (5.0, 6.25, and 8.33\,Hz) and report the best result.
We train OPT-125M~\cite{zhang2022opt} language models on top of the resulting units, following the same setup as for ZeroSyl above, except for SyllableLM, which is trained on two GPUs for 50\% more steps to account for the higher token rate.
For scaled comparison to SpidR~\cite{poli2025spidr}, we use results directly from the paper.

\section{Experiments: Syllabic system comparison}
\label{sec:experiments}

We perform a range of experiments to benchmark ZeroSyl's units.
In this section, we focus on systems that operate at the syllable level; comparisons against other systems are reserved for Section~\ref{sec:scaled}.
We therefore compare ZeroSyl against Sylber~\cite{cho2025sylber} and SyllableLM~\cite{baade2025syllablelm} when trained on 6k hours of Libri-Light.
This is a reasonable training set size enabling reliable comparisons without extensive training time.
Before describing language modeling results, we first perform intrinsic evaluations of
the predicted syllable boundaries and discovered syllabic clusters.

\begin{table}[t]
    \centering
    \caption{
        Boundary detection performance on the combined test sets of LibriSpeech.
        A boundary tolerance of 50\,ms is allowed.
        Pr: precision, Re: recall, R: R-value, OS: over segmentation.
    }
    \vspace*{-5pt}
    \label{tbl:boundary_results_compare}
    \renewcommand{\arraystretch}{0.9}
    \begin{tabularx}{\columnwidth}{@{}p{2.6cm}RRRRRRRR@{\,}}
        \toprule
        & \multicolumn{5}{c}{Boundary} & \multicolumn{3}{c}{Token} \\
        \cmidrule(l){2-6}\cmidrule(l){7-9}
        & Pr & Re & F1 & OS & R & Pr & Re & F1 \\
        \midrule
        SyllableLM 5.0\,Hz & \textbf{71} & 79 & \textbf{75} & 10 & \textbf{77} & \textbf{54} & 58 & \textbf{56} \\
        SyllableLM 6.25\,Hz & 62 & 87 & 72 & 41 & 59 & 44 & \textbf{59} & 50 \\
        SyllableLM 8.33\,Hz & 50 & \textbf{91} & 65  & 81 & 27 & 31 & 52 & 39 \\
        Sylber & 65 & 74 & 69 & 13 & 71 & 48 & 54 & 51 \\
        PromSeg & 62 & 65 & 64 & \hphantom{0}\textbf{4} & 69 & 40 & 41 & 41 \\
        ZeroSyl & 69 & 75 & 72 & 10 & 75 & 52 & 56 & 54 \\
        \bottomrule
    \end{tabularx}
    \vspace{-5pt}
\end{table}

\begin{table*}[b!]
    \centering
    \vspace*{-4pt}
    \caption{
        Syllable discovery and language modeling performance.
        Syllable discovery metrics are reported on the combined LibriSpeech test set.
        Language models are trained on 6k hours of Libri-Light and evaluated on the sLM21 development set.
    }
    \vspace*{-5pt}
    \label{tbl:comparative_discovry_and_lm_results}
    \renewcommand{\arraystretch}{0.9}
    \begin{tabularx}{\textwidth}{@{}l@{}RRRRRRRRRR@{}}
        \toprule
        
        &
        &
        &
        &
        \multicolumn{3}{c}{Syllable discovery (\%)} &
        \multicolumn{4}{c}{Language modeling (\%)}  \\
        
        \cmidrule(l){5-7}\cmidrule(l){8-11}
        
        &
        Vocab. size&
        Bitrate (bps) &
        Freq. (Hz) &
        Purity \hphantom{000} &
        Inverse purity &
        SNMI \hphantom{0} &
        sWUGGY (all) &
        sWUGGY (IV) &
        sBLIMP \hphantom{0}&
        tSC \hphantom{000} \\
        
       \midrule
       
        SyllableLM 6.25Hz  & 8k & 73 & 5.87 & 67.5 & 23.2 & 82.6 & 66.1 & 74.2 & 56.4 & 67.6   \\
        Sylber  & 10k & 53 & 4.46 & 73.5 & 28.4 & 83.5 & 66.0 & 74.7 & 59.1 & 65.8 \\
        ZeroSyl (uncollapsed sil.) & 10k & 58 & 4.55 & \textbf{80.5} & 20.0 & \textbf{89.5} & 67.0 & 76.3 & 58.6 & 67.5 \\
        ZeroSyl  & 9k & \textbf{52} & \textbf{4.35} & 79.9 & \textbf{32.0} & 88.9 & \textbf{68.0} & \textbf{78.6} & \textbf{60.5} & \textbf{68.1} \\
        
        \bottomrule
    \end{tabularx}
\end{table*}

\subsection{Boundary and token detection}

\textbf{Setup:}
We start by evaluating how closely the predicted boundaries match true syllables.
Ground truth syllable boundaries are obtained using phone-level forced alignments~\cite{mcauliffe2017mfa} with a rule-based method~\cite{gorman2013syllabify} that converts phone sequences to syllables.
In addition to precision, recall, and F1, boundaries are also evaluated using R-value and token scores.
R-value measures how close predicted syllable boundaries are to an ideal operating point with 100\% recall and 0\% over-segmentation~\cite{rasanen2009rvalue}; over-segmentation is the ratio of the number of predicted boundaries relative to the number of ground-truth boundaries~\cite{petek1996overseg}.
Token boundary scores (precision, recall, and F1) require both the start and end boundaries to be correct.
For all metrics, we allow a boundary tolerance of 50 ms on either side of a true boundary~\cite{cho2025sylber,baade2025syllablelm}.
Boundaries that border silences are not evaluated.
As in~\cite{cho2025sylber}, we allow for a constant time shift in the predicted boundaries to account for potential latency in the representations; 
these shifts are tuned for each system on development data.

\textbf{Results:}
Table~\ref{tbl:boundary_results_compare} presents the boundary prediction results on the combined test sets of LibriSpeech.
Overall, the 5.0\,Hz version of SyllableLM performs best on several of the metrics. However, ZeroSyl performs competitively despite not requiring any model training, achieving very similar token F1 scores (54\% vs.\ 56\%).
With an R-value of 75\% and a token F1-score of 54\%, ZeroSyl also outperforms Sylber (71\% and 51\%, respectively).
ZeroSyl maintains a low over-segmentation rate of 10\%,
substantially lower than the higher-frequency SyllableLM variants.
This indicates that our method avoids fragmenting syllables into smaller acoustic events.
We also include a prominence-based baseline (PromSeg) that uses the same layer, window size, and prominence as ZeroSyl, but with peak detection on the cosine distance between adjacent frames~\cite{pasad2023words}.
ZeroSyl outperforms this baseline on all metrics except over-segmentation.
This shows that the L2 norm provides a better signal for syllable boundaries.

\subsection{Syllable discovery}

\textbf{Setup:}
To evaluate the quality of the discovered syllabic units, we assess how well the clusters map to true syllable labels.
As in \cite{cho2025sylber}, we map predicted segments to ground-truth syllables with the most overlap.
High-quality clusters should exhibit 
high per-cluster purity,
where each cluster ID maps mostly to one syllable label, as well as high inverse purity,
where each syllable is found mostly in one cluster.
Syllable-normalized mutual information (SNMI) measures the percentage of uncertainty about the syllable label eliminated after observing the cluster~ID~\cite{hsu2021hubert}.
While smaller vocabularies are often used to evaluate syllabic clusters \cite{cho2024sdhubert, cho2025sylber, baade2025syllablelm}, we evaluate the systems using the actual vocabulary sizes employed in the downstream language model (8k--10k).
We compare system efficiency in terms of bitrate (using the entropic formulation~\cite{dunbar2022lessons}) and token frequency.
Evaluations are done on the LibriSpeech test set.

\textbf{Results:}
Table~\ref{tbl:comparative_discovry_and_lm_results} shows syllable discovery results in its middle columns.
By default, ZeroSyl collapses silences (Section~\ref{sec:silence_collapsing}), but here we also report scores without collapsing.
ZeroSyl outperforms all other systems on all purity and mutual-information metrics, while maintaining the lowest bitrate (52~\,bps).
Specifically, ZeroSyl achieves an SNMI of 88.9\%, higher than Sylber (83.5\%) and SyllableLM (82.6\%).
The rich semantic information in the later layers of WavLM Large likely boosts the quality of the representations compared to the base versions of data2vec~\cite{baevski2022data2vec} and SD-HuBERT~\cite{cho2024sdhubert} that are used as backbones in SyllableLM and Sylber, respectively.
Comparing the uncollapsed and collapsed versions of ZeroSyl shows that merging redundant silence tokens drastically improves inverse purity (from 20.0\% to 32.0\%) while maintaining high per-cluster purity and SNMI.

\subsection{Spoken language modeling}
\label{sec:spoken_language_modeling_comparative}

Our main question is whether syllabic units from the simple ZeroSyl approach (Sections~\ref{sec:boundaries} and~\ref{sec:silence_collapsing}) are useful for spoken language modeling.
We train language models on the syllabic units of ZeroSyl (Section~\ref{sec:lm_details}) and the other systems (Section~\ref{sec:baselines}), and probe the models' lexical, syntactic, and narrative capabilities.
All the benchmarks below rely on likelihood estimates from the language model, for which we use the mean log-likelihood per token to account for varying sequence lengths.

\textbf{Lexical task (sWUGGY):}
This task tests whether a model can distinguish real lexical items from phonologically plausible fakes.
The model processes pairs containing one actual word, such as ``manufacturer," and one matched non-word, like ``manifelturer," and must assign a higher likelihood to the real word.
We report metrics for \textit{all} pairs in the SLM21 development set~\cite{dunbar2021zrc}, and for \textit{in-vocabulary (IV)} pairs, where the real word is guaranteed to appear in the LibriSpeech vocabulary.

\textbf{Results:}
Table~\ref{tbl:comparative_discovry_and_lm_results} shows that ZeroSyl outperforms all the other systems on the sWUGGY benchmark: it
achieves the highest accuracy on both the general (68.0\%) and in-vocabulary (78.6\%) subsets.
The silence collapsing step in ZeroSyl also yields a clear improvement over the uncollapsed version.

\textbf{Syntactic task (sBLIMP):}
This benchmark evaluates a model’s grasp of sentence structure~\cite{dunbar2021zrc}.
Each test item consists of two spoken sentences: one is grammatically correct, while the other contains a single syntax error.
For example, the model compares ``What \textit{dentist} is Angela working with?" against ``What is Angela working with \textit{dentist}?" and must identify the valid sentence by assigning it a higher likelihood.

\textbf{Results:}
Table~\ref{tbl:comparative_discovry_and_lm_results} shows that ZeroSyl achieves an sBLIMP score of 60.5\%, higher than both SyllableLM (56.4\%) and Sylber (59.1\%).\footnote{We report all scores using mean per-token log-likelihoods, since this gave better sWUGGY and tSC scores for all systems, and since this is what is reported in~\cite{borsos2023audiolm, poli2025spidr}. However, when using unnormalized log-likelihoods, sBLIMP scores improve to 62.4\%, 60.8\%, 63.0\%, and 63.8\% for SyllableLM, Sylber, Zerosyl-uncollapsed, and ZeroSyl, respectively.}
Again, the collapsed version of ZeroSyl improves upon the one where silences are not collapsed.

\textbf{Narrative task:}
The Topic StoryCloze (tSC) benchmark evaluates long-range language modeling performance by measuring narrative coherence \cite{hassid2024twist}.
The model receives a four-sentence spoken prompt and must identify the original fifth sentence from a negative sample randomly selected from a different story.
This task tests whether the model can pick a logically consistent continuation in a spoken context.

\textbf{Results:}
As shown in Table~\ref{tbl:comparative_discovry_and_lm_results}, ZeroSyl achieves 68.0\% accuracy on the tSC task, outperforming all other systems.

Overall, ZeroSyl outperforms other syllabic units systems across the lexical, syntactic, and narrative benchmarks.
Given that ZeroSyl also has the lowest bitrate, these results show that the discovered units provide a more efficient representation for spoken language modeling than previous work.

\section{Experiments: Scaling behavior}
\label{sec:scaled}

Finally, we investigate the performance of ZeroSyl as training data scales from 600 to 6k to 60k hours of Libri-Light.
Here we compare mainly to SpidR~\cite{poli2025spidr}, because it represents a well-performing non-syllabic approach which is still simple (like ZeroSyl): SpidR uses $K$-means to  discretize features from a proposed SSL model.
However, unlike ZeroSyl, this system produces tokens close to the frame-level,\footnote{With $K=256$, the token rate is 26\,Hz and the bitrate is 205\,bps.}
resulting in a much higher bitrate than ZeroSyl.
We also include the text topline from~\cite{poli2025spidr} that uses byte-pair encoding (BPE)
on characters.
ZeroSyl, SpidR, and the topline all use the same language model~\cite{zhang2022opt}.

\begin{figure}[t]
  \centering
  \includegraphics[width=\linewidth]{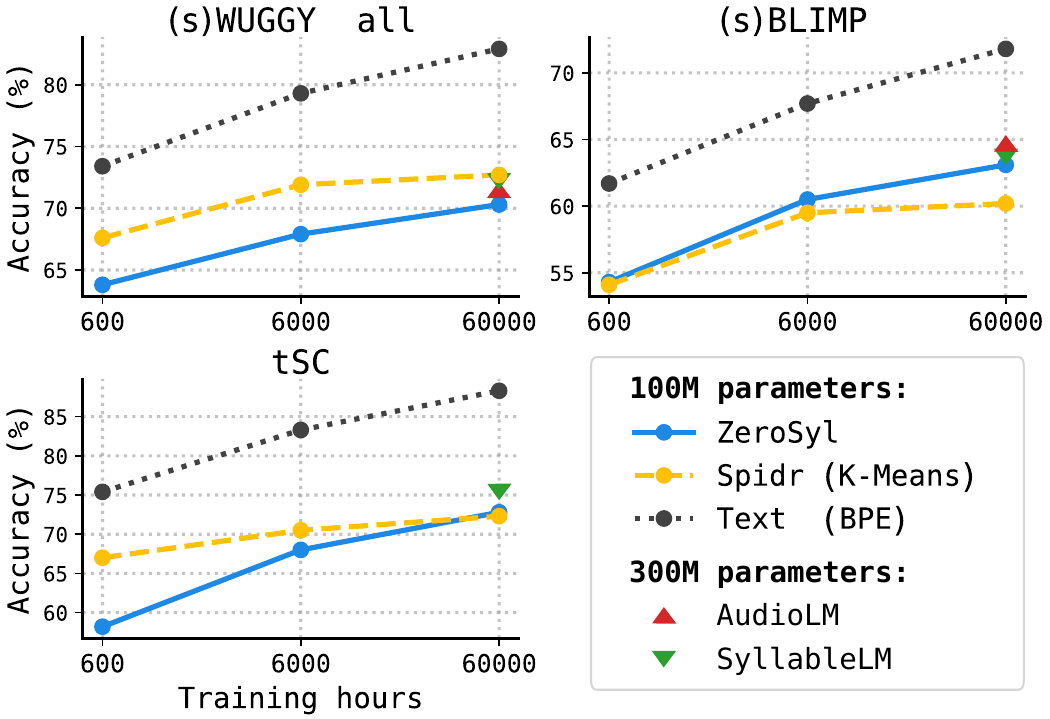}
  \vspace*{-15pt}
  \caption{
    Scaling behavior of ZeroSyl compared to speech (SpidR with $K$-means) and text (BPE) systems on the Libri-Light corpus.
  }
  \label{fig:scaling-results}
  \vspace*{-7.5pt}
\end{figure}

\textbf{Results:}
sWUGGY, sBLIMP, and tSC results are given in Figure~\ref{fig:scaling-results}.
In the sWUGGY task~(Figure~\ref{fig:scaling-results} top-left), ZeroSyl trails the fine-grained SpidR.
This reflects a trade-off: while SpidR units preserve the frame-level phonetic detail required in the lexical task, ZeroSyl compresses this information into syllabic units, sacrificing acoustic granularity.
However, this trade-off proves advantageous at the syntactic level:
in the sBLIMP evaluation~(Figure~\ref{fig:scaling-results} top-right), the performance of SpidR saturates as data scales, whereas ZeroSyl maintains a steep upward trajectory, outperforming SpidR at 6k and 60k hours.
On the tSC narrative task~(Figure~\ref{fig:scaling-results} bottom-left), ZeroSyl effectively matches SpidR at the 60k-hour mark, with a steeper upward trend.

Throughout this paper, we used the same 100M-parameter language model architecture~\cite{zhang2022opt} to allow for meaningful comparisons, and did not consider the considerably larger sizes also occasionally used~\cite{borsos2023audiolm,baade2025syllablelm}.
But we see in Figure~\ref{fig:scaling-results} that the performance gains of the 300M-parameter AudioLM~\cite{borsos2023audiolm} and SyllableLM~\cite{baade2025syllablelm} systems are marginal considering the significant increase in computational cost.
Interestingly, even with these bigger models, a single system does not perform best on all three tasks.\footnote{AudioLM did not evaluate the tSC benchmark or release models.}
Returning to our main question of whether a simple syllable-based approach is possible, it is clear that even compared to these much larger models, ZeroSyl provides competitive performance without intricate training procedures.

\section{Conclusion}

In this work, we introduced ZeroSyl, a training-free framework for discovering syllabic units directly from unlabelled speech.
By leveraging feature norms from a frozen SSL model (WavLM Large), ZeroSyl eliminates the complex multi-stage methods required by prior state-of-the-art syllabic tokenizers.

ZeroSyl outperforms existing syllabic tokenizers across lexical, syntactic, and narrative benchmarks when trained on 6k hours of speech.
Scaling experiments reveal that while finer-grained units retain an advantage for lexical tasks, ZeroSyl's syllabic units are more effective for modeling long-range dependencies, showing a steep upward trajectory in syntactic performance as data increases.
Ultimately, ZeroSyl demonstrates that high-quality spoken language modeling does not necessarily require increasingly complex tokenizers, but rather a deeper understanding of the signals already present in SSL features.

Future work should investigate why and how the L2 norm signal of SSL representations provides implicit positional encodings of syllables.
While ZeroSyl's low rate encoding benefits syntactic and narrative modeling, the resulting syllabic units might be detrimental for encoding lexical items, especially if they are rare or unseen.
This should be investigated.

\section{Acknowledgements}

Nicol Visser and Simon Malan are funded through the Google PhD Fellowship Program.

\section{Generative AI use disclosure}

Generative AI tools were utilized exclusively for linguistic refinement, copy-editing, and polishing of the author-drafted text to improve clarity. No scientific content, data analysis, or significant portions of the manuscript were generated with these tools; the authors retain full responsibility for the accuracy, originality, and intellectual integrity of the work.

\bibliographystyle{IEEEtran}
\bibliography{mybib}

\end{document}